\documentclass[letterpaper, 10 pt, conference]{ieeeconf}
\pdfoutput=1

\IEEEoverridecommandlockouts    

\usepackage{booktabs}
\usepackage[pdftex]{graphicx} 
\usepackage{gensymb}
\usepackage{amsmath}
\usepackage{hyperref}
\usepackage{amssymb}
\usepackage{xcolor}

\title{Multimodel Sensor Fusion for Learning Rich Models for Interacting Soft Robots}

\author{Thomas George Thuruthel$^{1}$, Fumiya Iida$^{1}$
\thanks{
This work was supported by the SHERO project, a Future and Emerging Technologies (FET) programme of the European Commission (grant agreement ID 828818).} 
\thanks{$^{1}$ The Bio-Inspired Robotics Lab, Department of Engineering, University of Cambridge, UK.} 
        \thanks{Source code and data for this paper can be found in : \url{https://github.com/tomraven1/Soft_Deep_Models}}%
}

\begin{document}

\maketitle

\begin{abstract}

Soft robots are typically approximated as low-dimensional systems, especially when learning-based methods are used. This leads to models that are limited in their capability to predict the large number of deformation modes and interactions that a soft robot can have. In this work, we present a deep-learning methodology to learn high-dimensional visual models of a soft robot combining multimodal sensorimotor information. The models are learned in an end-to-end fashion, thereby requiring no intermediate sensor processing or grounding of data. The capabilities and advantages of such a modelling approach are shown on a soft anthropomorphic finger with embedded soft sensors. We also show that how such an approach can be extended to develop higher level cognitive functions like identification of the \textit{self} and the external environment and acquiring object manipulation skills. This work is a step towards the integration of soft robotics and developmental robotics architectures to create the next generation of intelligent soft robots. 

\end{abstract}

\section{INTRODUCTION}

Soft robots are characterized by a large number of passive Degrees of Freedom (DoF). These systems can be analytically modelled using continuos infinite-dimensional systems or using finite-dimensional approximations \cite{della2021model,armanini2021soft}. Infinite-dimensional models are solved by finding an approximate solution to a partial differential equation. The accuracy of the model can be increased by refining the discretization size. Examples include continuum models \cite{trivedi2008geometrically,rucker2011statics,renda2014dynamic} and finite element models \cite{duriez2013control,goury2018fast}. Finite-dimensional models rely on some sort of geometric assumptions. Hence, they have limited accuracy but are computationally cheaper. Examples include constant curvature \cite{webster2010design} and piecewise constant curvature models \cite{della2020improved}. However, developing analytical models for a soft robot is challenging because of the complexity of the system. Hence, learning-based modelling approaches are becoming popular to obtain accurate models at lower computational costs without any prior assumptions on the underlying physics \cite{chin2020machine,george2018control,kim2021review}. 

Data-driven modelling of soft bodied robots have been used for state estimation, kinematic and dynamic control. State estimation models try to estimate state information from raw sensor data. Typically, these works focus on obtaining interoceptive and exteroceptive information from nonlinear embedded sensors \cite{han2018use,wall2017method,thuruthel2019soft}. Kinematic models develop steady-state mappings from the actuator input to the robot geometry \cite{giorelli2013feed,yip2016model,george2017learning}. First order \cite{george2020first} and second order dynamical models \cite{thuruthel2017learning,gillespie2018learning,bruder2019modeling} can be learned to develop controllers for dynamic and dexterous tasks.

\begin{figure}[t]
\centering
\includegraphics[width=\linewidth]{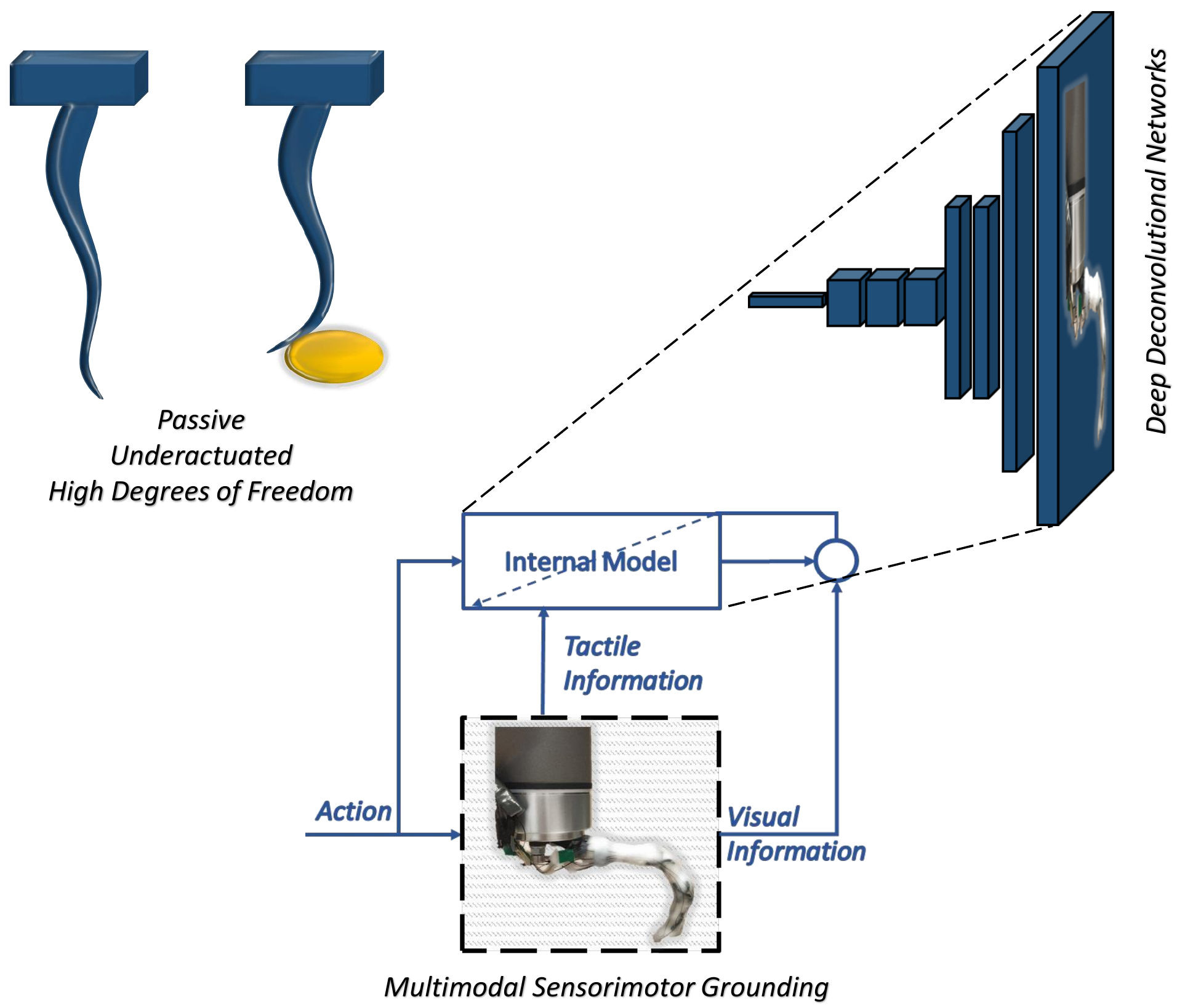}
\caption{Soft robots are characterized by large passive Degrees of Freedom that makes them interact with the environment in multiple configurations. Estimating this high-dimensional state of the robot requires integration of rich sensorimotor data.}
\label{fig:fig1}
\end{figure}

\begin{figure*}[t]
\centering
\includegraphics[width=\linewidth]{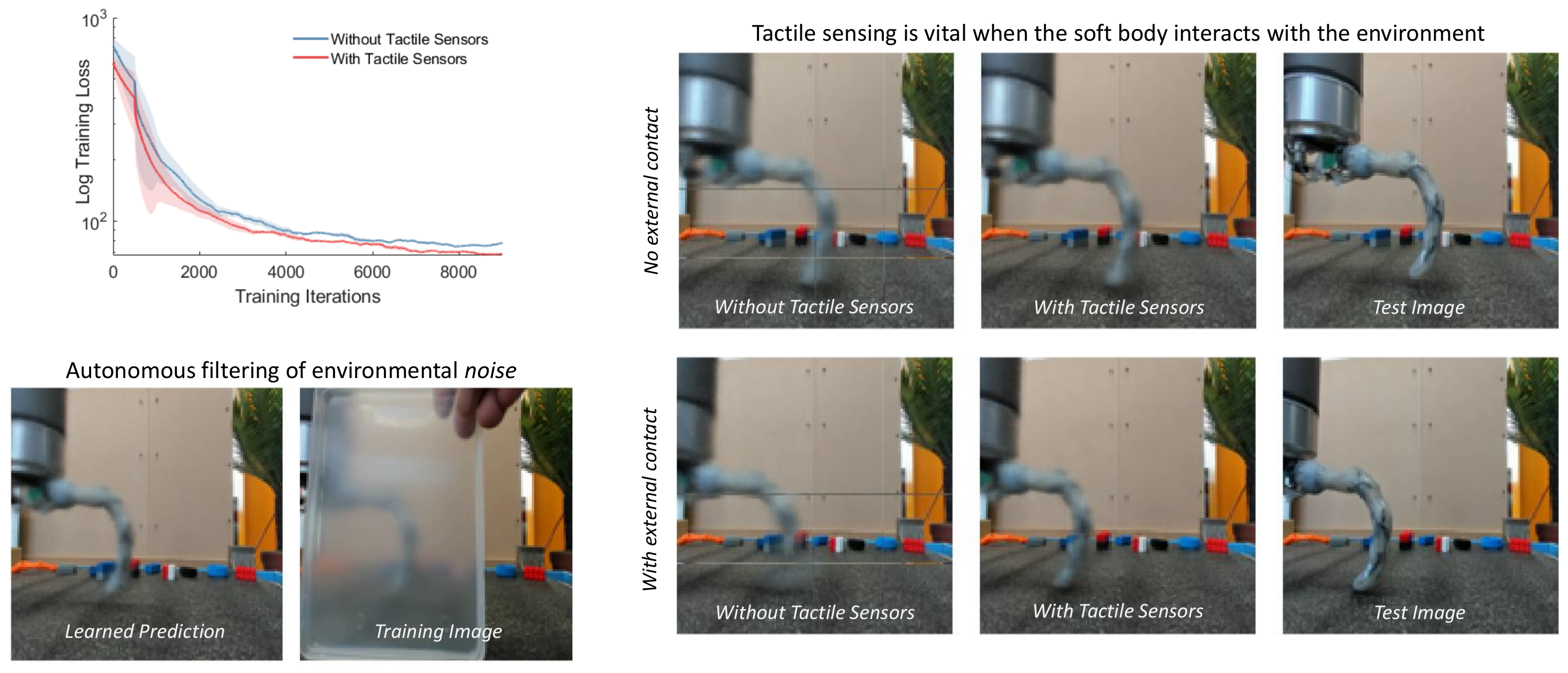}
\caption{Top left: Training performance of the deep network. Top Right: Tactile data is essential for accurate predictions of the body schema when a passive soft robot interacts with the environment. Bottom left: The body schema architecture automatically filters out uncorrelated image states, even in the training set.  }
\label{fig:res-passive}
\end{figure*}

Unlike analytical formulations, learning-based approaches tend to learn only low-dimensional models, be it for state estimation, kinematics or dynamic control. This is because learning-based approaches high-dimensional models require large amount of data to be trained and do not provide additional advantages when the soft robot does not perform complex interactions with the environment. Even though a soft robot has large degrees of freedom, most of these passive DoFs do not get excited in contact-free motions. Modelling soft robots that can have unconstrained interactions with the environment is a challenging problem. Now the effective DoFs can increase significantly based on the interaction scenario (Figure \ref{fig:fig1}). To reconstruct the state of the interacting soft robot, an internal model of the robot is required that can combine contact information from tactile sensors and motion input to the robot (for a work that presents a solution to the inverse of this problem, please refer to \cite{zhang2018vision}). There have been few FEM-based methods that have looked into the soft robotic interactions. The most relevant is a model-based approach to estimate force and deformation estimate of a soft body using embedded sensors \cite{navarro2020model}. Soter et. al. presented one of the first works on developing high-dimensional state estimation models for soft robots \cite{soter2018bodily}. They presented a deep learning approach for predicting the high-dimensional visual state of a soft robot using only embedded strain sensor data. However, interactions with an unstructured environment were still not considered in their study, which in when such high-dimensional models become relevant. 

This work presents a learning-based approach for developing high-dimensional internal models (analogous to FEM and continuum models in model-based techniques) for soft robots incorporating deformations through external interactions. We use touch and vision data to develop sensorimotor models that are grounded on the raw sensor data (See Figure \ref{fig:fig1}). The key idea is to learn predictive models of visual sensory response using just efferent (commanded) action data and afferent (observed) tactile information. Such a method allows us to develop rich internal models of a soft robot, in a data-driven manner, while incorporating deformation models using embedded sensors. The modelling approach is demonstrated on a soft anthropomorphic finger with embedded soft strain sensors. We demonstrate how accurate rich visual models can be learned using deep learning methods. The importance of tactile sensors for self-modelling in soft robotics is shown through our experiments. Finally, we demonstrate techniques for acquiring object manipulation skills which leads to interesting emergent cognitive skills, vital for an intelligent system.

\subsubsection{Related work in developmental robotics}

Humans, over the course of their development, learn to perceive our body in space, sense the location of our limbs, identify and categorize external agents, and develop models to interact with them \cite{bremner2012multisensory}. The field of developmental robotics investigates how such models are learned and represented in the brain to develop next generation intelligent robots \cite{nguyen2021sensorimotor}. Here, we briefly enumerate some relevant and related works for interested readers. Finn. et. al. presented an unsupervised learning algorithm for predicting object interactions using just motion and vision data \cite{finn2016unsupervised}. Integration of sensory information (touch, vision, and proprioception) for adaptive learning of the body schema was presented by Lanillos et. al. \cite{lanillos2018adaptive}. A similar work was shown by Laflaquiere et. al. \cite{laflaquiere2019self}. The role of these sensory predictive models in the emergence of cognitive skills was demonstrated by Lang. et. al., where in, they showed a sense of agency and the capability to maintain an enhanced internal visual representation of the world \cite{lang2018deep}. More recently, there have been several works on models for self-supervised self-awareness or self-recognition in robotic systems \cite{almeida2021my,hoffmann2021robot}. 

\section{Interactions with a stationary environment}

\subsubsection{Methodology}
In this section, we describe the architecture for developing rich body schema of soft robots by sensorimotor integration. For all our experiments, we use a soft anthropomorphic finger mounted on a Universal Robots UR5 manipulator (Figure \ref{fig:fig1}). The finger is made of a 3D printed skeleton joined by artificial ligaments and covered with a silicon skin. The finger has six soft sensors embedded in a random configuration (Refer to \cite{thuruthel2020drift} for more details on the hardware). The raw sensor resistances are measured using a NI-USB 6212 data acquisition board. These sensors provide the only contact information to the system. Being a soft robotic sensor, they exhibit temporal non-linearities like drift and hysteresis, which makes them difficult to model (we later show, however, that these non-linearities can be used a memory reservoir for short term predictive tasks). A Logitech BRIO webcam is mounted at a fixed location, directed at the system. MATLAB programming interface is used to control the UR5 robot in three axes (X,Y,Z). The raw tactile sensor data and camera images are stored at sample rate of 33 Hz. The RGB image is downsampled to a size of 128x128x3 to reduce training time.  

Our aim is to develop an internal visual body schema of the soft robot using the three-dimensional motion information and strain sensor data. A 25 layer deep neural network is used to learn this mapping from the nine dimensional input space to the high-dimensional image space. The efferent action signals and afferent tactile sensor data is combined and transformed to a 3x3x1 image-like representation. This input goes through a series of convolutional layers and transposed convolutional layers sandwiched by Rectified Linear Unit (ReLU) activation functions. The output from the deep network is the predicted visual scene (Refer to the shared source code for more details). A total of 77 batches of sample points are obtained. Each batch involves the soft finger randomly moving and interacting with the fixed environment. Visual 'noise' is added to the system to 1) prevent overfitting, 2) observe its effect on the learned internal representation. Each batch contains 5000 sample points, totalling to 385,000 data points. One batch is used for as the test set. The deep network is developed and trained in the MATLAB programming environment on an Nvidia GeForce RTX 3080 10Gb GDDR6X. 

\subsubsection{Results}

The training process of the deep network with and without tactile sensor data is shown in Figure \ref{fig:res-passive}. Even without information from the strain sensors, the network can predict the expected visual data very well as seen from the learning curve. This is because most of the states in the visual data are static or dependent only on the motion data. However, tactile information is vital when the finger interacts with the environment (See right side of Figure \ref{fig:res-passive}). In this case, the state of the soft finger cannot be estimated only by the current efferent action signal (The deformation of the finger depends on the time history of actions). This ambiguity can be resolved with the tactile sensor data. As the soft body becomes more complex and has more passive interactions, the relevance of tactile information increases. 

\begin{figure}[t]
\centering
\includegraphics[width=0.8\linewidth]{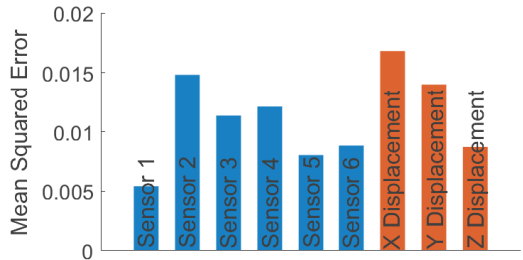}
\caption{The role of each input element towards body-schema prediction error. }
\label{fig:input_dep}
\end{figure}

The proposed body schema architecture also leads to interesting emergent behaviour. As the only input source provided to the network is the motion data and the strain sensor data, any variations in the visual scene caused by external factor automatically gets filtered out (See figure \ref{fig:res-passive}). Hence, a straightforward usage of such model is as a replacement of visual sensors when there are occlusions. For example, a soft robotic catheter with embedded sensors can be trained using the proposed architecture. Once deployed inside the body, a rich visual representation of the soft robot can still be obtained, provided there are sufficient strain sensors to capture the required deformation information. Moreover, we can also foresee how such architectures can lead to the emergence of self-awareness in an unsupervised manner. If the background image was not static (the robot was mounted on a mobile platform or the camera is mounted on the robot itself), then with sufficient data and learning, all states (pixels in the image) that are not part of the body (self) will average out, leading to clean segmentation of the body from the environment. This phenomenon is further investigated in the next section.

\begin{figure}[t]
\centering
\includegraphics[width=0.8\linewidth]{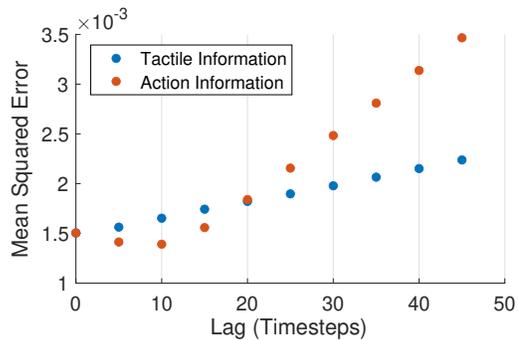}
\caption{Prediction error dependence on the lag of efferent action signal and afferent tactile signal. The prediction errors indicate that the actual motions lag behind the efferent action signals.}
\label{fig:lag_depen}
\end{figure}

The role of each input element towards the overall body schema prediction is shown qualitatively in Figure \ref{fig:input_dep}. This is done by removing the information from each element and testing its performance on the pre-trained network. As expected, for our scenario, certain action variables have more importance over the other because of the viewing angle. It can also be seen that certain strain sensors have more influence towards the predictions than the others. This could be because they have a higher response to the specific interaction scenario. Note that once the network is trained, the tactile sensor data has comparable importance to the motion data, indicating that their information is fused to create predictions. It is to be investigated if this is still the case for a redundant sensor network. The effect of information lag on the predictions is shown in Figure \ref{fig:lag_depen}. This is again tested on the network after training on the original data. As expected, when the efferent tactile sensor data is delayed, the predictions increase in proportion to increasing lag. However, when the efferent action signal is delayed, the predictions actually improve. A lag of 10 time steps (around 0.3 seconds) seems to provide the best prediction results. We can hypothesize that this is because the actual motion of the robot will always lag behind the efferent action signals due to the time spent in processing and converting the digital signals to electrical signals, and the dynamics of the robot itself.

\begin{figure*}[t]
\centering
\includegraphics[width=0.6\linewidth]{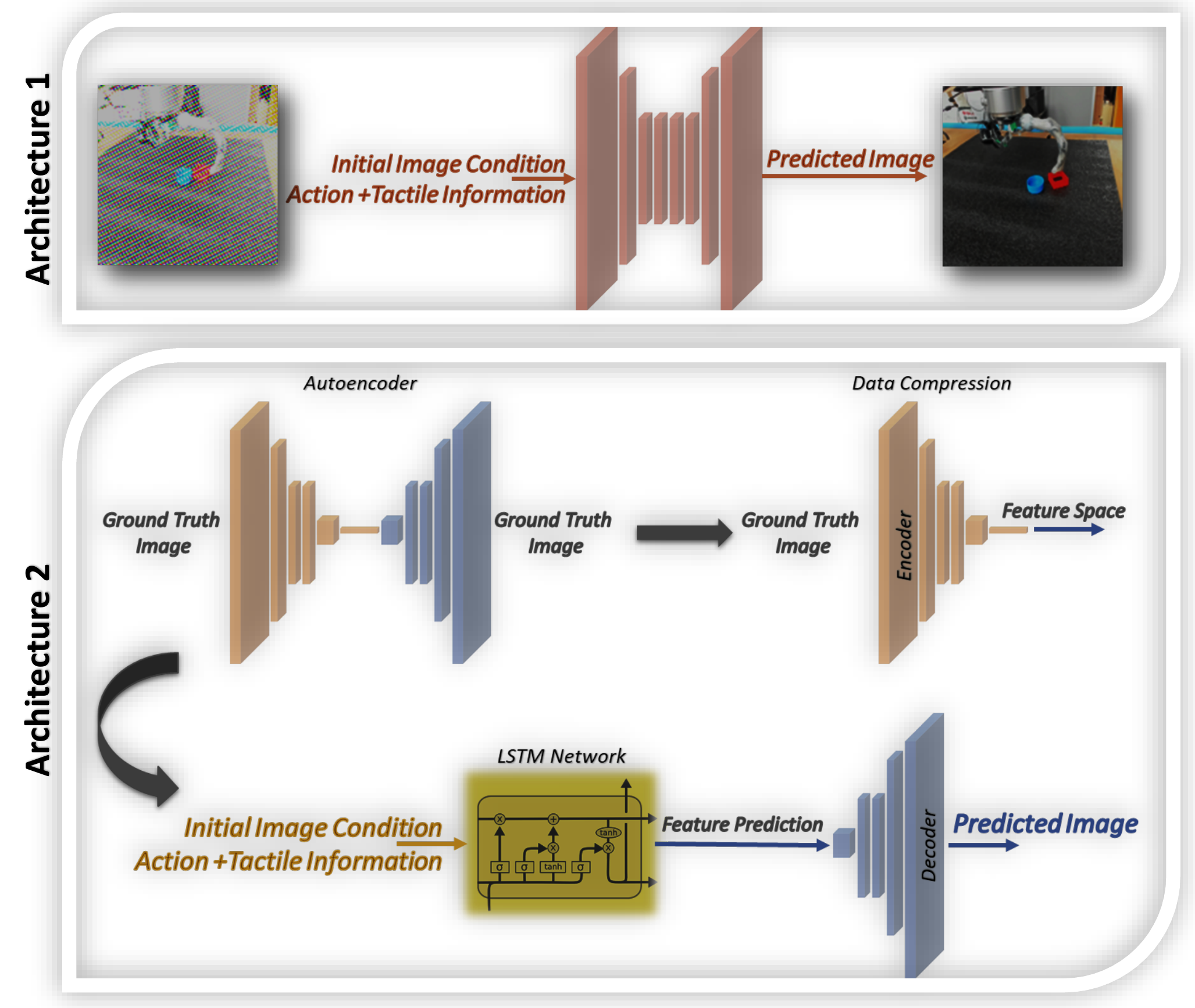}
\caption{Two learning architectures investigated in this work. The first architecture uses a deep static network to predict the expected visual scene from the initial conditions and current actions and sensor state. The second architecture uses a two-stage learning process so that the same data mapping is done using a deep recurrent neural network. }
\label{fig:twoarchi}
\end{figure*}

\section{Interactions with a non-stationary environment}

In this section, we augment the previously described architecture in order to make predictive models of controllable objects in the environment (Figure \ref{fig:twoarchi}). Such models are an essential precursor for developing learning-based controller for object manipulator. In principle, to develop such predictive models, a network would need to know the initial condition of the scene (location and type of objects, position of hand, etc), the set of actions and observed tactile signals and internal physics model of the object and the finger. We assume that contact information is essential to estimate the state of the object (The scenario can be pictured as a manipulation of objects inside a concealed box after looking inside it in the beginning).  

\subsubsection{Methodology}

As the problem is highly complex and requires extensive data collection and a rich tactile sensory system to be truly generalizable, we restrict our study to a selected few scenarios. A set of 11 objects are selected and placed in the scene randomly (Figure \ref{fig:res_example}). The robot is commanded to move randomly inside a fixed workspace, interacting with the object. The visual data, tactile sensor data and efferent motion signals are recorded. Fifty-six batches of different scenarios are created, each lasting 1000 sample points. No validation set is used for this scenario.   

Two learning architectures are investigated (See Figure \ref{fig:twoarchi}). The first architecture is a simple, memory-less, deep network. The initial fixed visual scene is appended with the motion and tactile data (by adding this information uniformly across the image space) and fed to a 26-layer deep network similar to the architecture described in the previous section. The main difference here is that the input to the network has the same dimensions as the output (128x128x3). In principle, the predictive capability of this architecture has to be short-sighted, as it does not have any way to update the states of the objects once it has been moved and the contact broken. 

The second architecture adds recurrent connections to provide a memory reservoir to the network. As the complexity of a deep convolutional network would increase exponentially with the addition of recurrent connections, we construct a two-stage learning process (Figure \ref{fig:twoarchi}). First, the ground truth image data is compressed to a low-dimensional feature space using autoencoders. A 12-layer deep network with convolutional layers and a 60 unit LSTM layer is then used to predict this feature variables using the augmented inputs (same as one used in the first architecture). The predicted features are then fed into the decoder of the previously trained autoencoder to get back the predicted images.   

\begin{figure}[t]
\centering
\includegraphics[width=\linewidth]{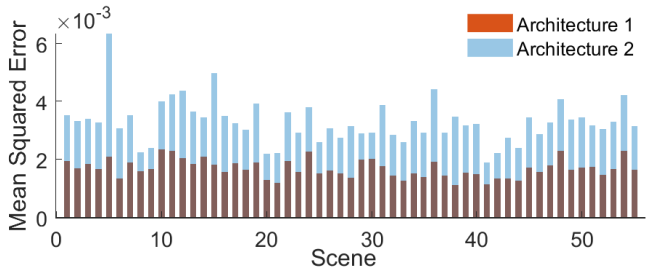}
\caption{Prediction error comparisons for the two architectures proposed in Figure \ref{fig:twoarchi}. }
\label{fig:comparison}
\end{figure}

\begin{figure}[t]
\centering
\includegraphics[width=\linewidth]{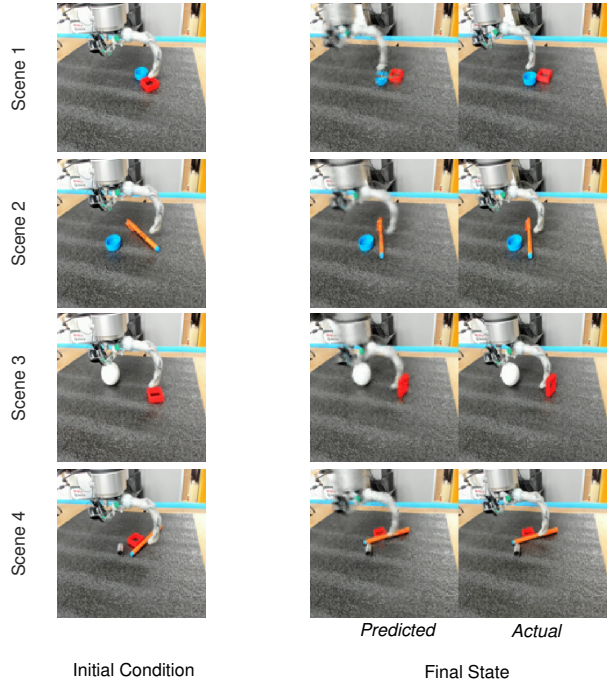}
\caption{Examples of predictions made by the model (Architecture 1) in different scenarios. The initial scene image and the subsequent actions and sensor states are used to make the image predictions on the right. }
\label{fig:res-example}
\end{figure}

\subsubsection{Results}

Figure \ref{fig:comparison} compares the prediction errors for the two architectures presented before. Remarkably, the static network performs better at this, essentially, time-series prediction task, when compared to the recurrent network. There are two factors that can contribute to this. First, the performance of the dynamic network is significantly affected because of the two-stage learning process. Compressing the original image to a low-dimensional feature spaces can lead to loss of information and the smoothness of the mapping from the inputs to the outputs. Second, the static network performs strikingly better because of the drift and hysteresis in the sensors. If the sensor responses go back to their initial conditions after an interaction, there is no information that the network can use to know the state of the displaced objects. This is not necessarily an advantage of these nonlinear sensors. 

Examples of the predictions made by the network learned from Architecture 1 are shown in Figure \ref{fig:res-example}. The initial condition of the scene is provided to the network as an image as shown in the left-hand side of the figure. Note that the architecture is static and similar to the one described in the previous section, and the variable information that the network receives (the nine-dimensional sensor and action data) is the same. Unlike the previous case, the changes in the external environment can be predicted very accurate now. This implies that the effect on the environment is directly correlated to the motion and sensor information which the deep network is able to capture.

\begin{figure*}[t]
\centering
\includegraphics[width=0.6\linewidth]{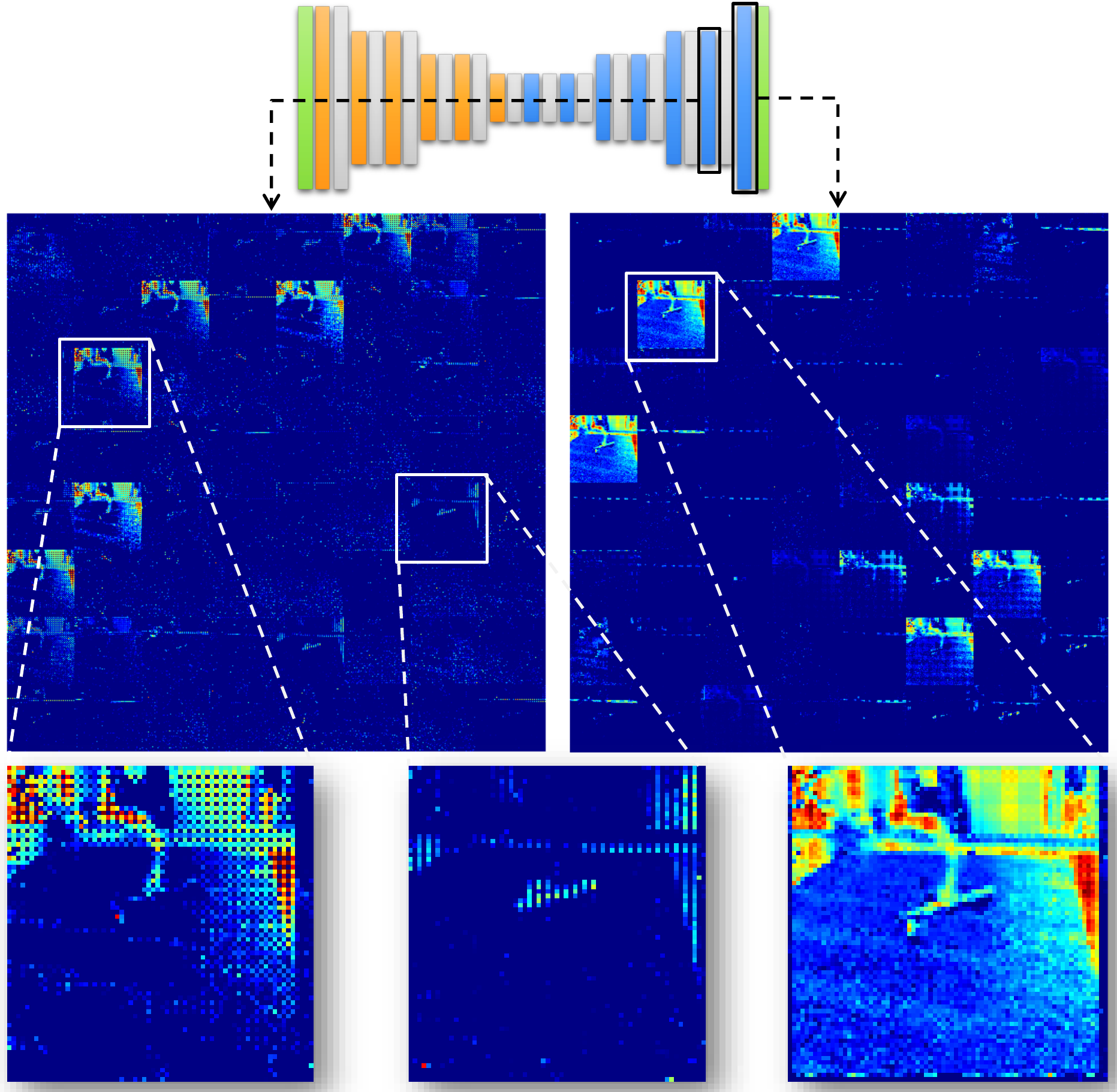}
\caption{Latent space visualization of the penultimate (left) and last (right) transposed convolutional layers. Self-segmentation of the body and the object can be observed in the penultimate latent space, which is later combined in the final layer.}
\label{fig:latent1}
\end{figure*}

There are, however, fundamental differences between the observed changes in the self and the changes in the manipulatable objects in the environment. The mapping between the inputs and the observed changes in the body is continuos, whereas for an external object, it is discrete and conditional. Hence, an emergent segmentation (division) between the two can be expected. Evidences of such can indeed be seen in the latent space of the learned deep network (See figure \ref{fig:latent1}). The latent space images are obtained by computing the activations of internal layers of the deep network. In the penultimate transposed convolutional layer of the network, a clear segmentation of the object (a pen, in this case) and a soft segmentation of the soft robot can be seen. These are later combined in the last layer. Such emergent segmentation can pave the way for a completely unsupervised identification of the self, the environment, and the objects in the environment that can be altered.  

\section{CONCLUSION}

This work presents a deep learning architecture for learning high-dimensional visual models of a soft robot combining multimodal sensorimotor information. The key idea is to learn predictive models of visual sensory response using just efferent action data and afferent tactile data, thereby obtaining a model that can essentially provide a mental image of the soft robot. As the model is learned in an end-to-end manner, the user does not need to further process or label the raw sensor data. The complexity of the underlying model is governed by the camera resolution and viewing angle. Stereo vision can be easily combined to the proposed architecture for more complex systems and interactions.  We show how this self-supervised learning approach can still lead to clear segmentation of the robot body from the environment and the identification and modelling of manipulatable objects. The importance of tactile sensing in the perception of the self, especially in a soft bodied system, is shown through our simple setup. 

Immediate applications of such high-dimensional models include the shape control of a soft robot and learning to manipulate unknown objects. As the models predict the expected sensory data itself, it can also be used as a substitute for visual sensors, in the presence of occlusions. As we need tactile sensor data to make predictions, such models cannot be used to develop predictive controllers for object manipulation. Hence, one of the future works include the development of predictive models of the tactile sensors using visuomotor information. Another scope for improvement is to develop and learn deep networks with recurrent connections directly without data compression.








\bibliographystyle{IEEEtran}
\bibliography{IEEEexample}

\end{document}